\PassOptionsToPackage{unicode}{hyperref}
\PassOptionsToPackage{hyphens}{url}
\documentclass[
]{article}
\usepackage{xcolor}
\usepackage{amsmath,amssymb}
\usepackage{iftex}
\ifPDFTeX
  \usepackage[T1]{fontenc}
  \usepackage[utf8]{inputenc}
  \usepackage{textcomp} % provide euro and other symbols
\else % if luatex or xetex
  \usepackage{unicode-math} % this also loads fontspec
  \defaultfontfeatures{Scale=MatchLowercase}
  \defaultfontfeatures[\rmfamily]{Ligatures=TeX,Scale=1}
\fi
\usepackage{lmodern}
\ifPDFTeX\else
\fi
\IfFileExists{upquote.sty}{\usepackage{upquote}}{}
\IfFileExists{microtype.sty}{% use microtype if available
  \usepackage[]{microtype}
  \UseMicrotypeSet[protrusion]{basicmath} % disable protrusion for tt fonts
}{}
\makeatletter
\@ifundefined{KOMAClassName}{% if non-KOMA class
  \IfFileExists{parskip.sty}{%
    \usepackage{parskip}
  }{% else
    \setlength{\parindent}{0pt}
    \setlength{\parskip}{6pt plus 2pt minus 1pt}}
}{% if KOMA class
  \KOMAoptions{parskip=half}}
\makeatother
\usepackage{longtable,booktabs,array}
\usepackage{calc} % for calculating minipage widths
\usepackage{etoolbox}
\makeatletter
\patchcmd\longtable{\par}{\if@noskipsec\mbox{}\fi\par}{}{}
\makeatother
\IfFileExists{footnotehyper.sty}{\usepackage{footnotehyper}}{\usepackage{footnote}}
\makesavenoteenv{longtable}
\usepackage{bookmark}
\IfFileExists{xurl.sty}{\usepackage{xurl}}{} % add URL line breaks if available
\hypersetup{
  hidelinks,
  pdfcreator={LaTeX via pandoc}}

\author{}
\date{}

\begin{document}

EARS: Efficient Adaptive Rejection Sampling for Accelerating Speculative
Decoding in Large Language Models

\textbf{Authors:} Chendong Sun, Ali Mao, Lei Xu, mingmin Chen

\textbf{Abstract:}\\
Speculative Decoding is a prominent technique for accelerating the
autoregressive inference of large language models (LLMs) by employing a
fast draft model to propose candidate token sequences and a large target
model to verify them in parallel. However, its core component---the
rejection sampling mechanism---relies on a fixed, context-independent
random threshold. This leads to a significant ``random rejection''
problem in high-uncertainty generation scenarios, where plausible
candidate tokens are frequently rejected due to random chance,
undermining inference efficiency. This paper introduces
\textbf{Efficient Adaptive Rejection Sampling (EARS)}, a novel method
that dynamically adjusts the acceptance threshold by incorporating the
target model's own predictive uncertainty, measured as $1 - \max(P_{\text{target}})$. By introducing a tolerance term
proportional to this uncertainty, EARS intelligently relaxes the
acceptance criterion when the model is uncertain, effectively reducing
random rejections while maintaining strict standards when the model is
confident. Experiments on creative writing and open-domain QA tasks
demonstrate that EARS significantly enhances the efficiency of
speculative decoding, achieving up to an \textbf{18.12\% increase in
throughput} with a negligible \textbf{0.84\% accuracy drop} on the GSM8K
benchmark. The method requires no modifications to model architectures
and can be seamlessly integrated into existing speculative decoding
frameworks.


\begin{center}\rule{0.5\linewidth}{0.5pt}\end{center}

\subsubsection{1. Introduction}\label{1-introduction}

Autoregressive inference in large language models (LLMs) suffers from
high latency due to its sequential nature. Speculative Decoding [1, 2] has emerged as
an effective acceleration paradigm, utilizing a fast
draft model to predict several future tokens, which are then verified in
parallel by a larger target model. The standard implementation employs a
rejection sampling-based verification mechanism: a candidate token from
the draft model is accepted if deemed sufficiently likely by the target
model; otherwise, it and all subsequent draft tokens are rejected, and
the target model regenerates from that point.

While effective in deterministic (temperature=0) or low-uncertainty
scenarios, we observe its efficiency degrades markedly in creative,
open-ended tasks (temperature $>0$). The root cause lies in the
traditional rejection sampling rule, which uses a uniformly distributed
random number $U \sim \text{Uniform}(0,1)$ as a fixed threshold.
This threshold completely ignores the intrinsic uncertainty of the
target model's predictive distribution at different generation steps.
When the model has multiple plausible next tokens (i.e., high entropy),
a reasonable but non-top candidate proposed by the draft model may have
its acceptance ratio $R = P_{\text{target}} / P_{\text{draft}}$ fall slightly
below $U$ merely due to random fluctuation. This ``random
rejection''---a rejection not based on candidate quality---wastes the
computation spent on verifying subsequent draft tokens and severely
limits the acceleration potential of speculative decoding in
high-uncertainty scenarios.

To address this, we propose \textbf{Efficient Adaptive Rejection
Sampling (EARS)}. The core idea is to make the acceptance threshold
aware of the target model's current predictive confidence. Specifically,
we define the model's uncertainty at a given position as
$\text{Uncertainty} = 1 - \max(P_{\text{target}})$ and compute
a dynamic tolerance $\text{Tolerance} = \beta \cdot \text{Uncertainty}$. The acceptance
condition is modified to $P_{\text{target}} / P_{\text{draft}} \geq U - \text{Tolerance}$.
This adjustment effectively lowers the threshold when the model is
highly uncertain (high $\text{Uncertainty}$), thereby sparing
plausible candidates that fall just below the original threshold,
reducing random rejections, and increasing the average accepted draft
length and overall throughput.

The main contributions of this paper are:

\begin{enumerate}
\def\labelenumi{\arabic{enumi}.}
\item
  We formalize the ``random rejection'' problem inherent in the
  traditional rejection sampling mechanism of speculative decoding for
  high-uncertainty generation.
\item
  We propose EARS, a simple yet effective adaptive rejection sampling
  algorithm that mitigates this problem by dynamically adjusting the
  acceptance threshold based on the target model's uncertainty.
\item
  We demonstrate how EARS can be integrated efficiently with minimal
  overhead via engineering optimizations like pre-computation and
  delayed lookup.
\item
  Our experiments on diverse tasks show that EARS significantly improves
  inference throughput (+18.12\%) while maintaining high output quality
  (accuracy drop $<$ 0.84\%).
\end{enumerate}

\subsubsection{2. Related Work}\label{2-related-work}

\textbf{Speculative Decoding:} The fundamental framework was introduced
concurrently by Leviathan et al. [1] and Chen et al. [2].
Subsequent works have optimized it by training better draft models
[3], designing multi-draft strategies [4], and improving the
compensation generation mechanism post-verification [5]. However,
these predominantly retain the original context-independent rejection
sampling rule.

\textbf{Sampling \& Decoding Strategies:} Standard strategies include
greedy decoding, beam search, and stochastic methods like top-k and
top-p (nucleus) sampling [6]. These focus on selecting tokens from a
distribution, whereas rejection sampling in speculative decoding focuses
on \emph{verifying} draft tokens. Our work improves this verification
rule.

\textbf{Adaptive Computation:} Some research explores dynamically
adjusting model computation based on input difficulty [7]. EARS
shares a similar spirit but operates at a different level: we adaptively
adjust the leniency of the ``accept draft'' decision based on the
model's real-time uncertainty, rather than altering the model's
computational graph.

\subsubsection{3. Method: Efficient Adaptive Rejection Sampling
(EARS)}\label{3-method-efficient-adaptive-rejection-sampling-ears}

\paragraph{3.1. Problem Formalization}\label{31-problem-formalization}

Let the target model be $M_t$ and the draft model be $M_d$. At each step of speculative decoding,
$M_d$ autoregressively generates a candidate sequence $\{x_1, ..., x_\gamma\}$ of length $\gamma$.
$M_t$ computes in parallel the conditional probability $P_t(x_i)$ for each $x_i$ given the true
prefix context, as well as the full distribution over the vocabulary.
$M_d$ also provides its generation probability $P_d(x_i)$.

The traditional rejection sampling rule is: for $i = 1$ to $\gamma$, sample $U_i \sim \text{Uniform}(0,1)$. If
$R_i = P_t(x_i) / P_d(x_i) \geq U_i$,
accept $x_i$ and continue; otherwise, reject $x_i$ and
all $x_{j>i}$, and have $M_t$ generate subsequent tokens autoregressively starting from
position $i$.

\textbf{The ``Random Rejection'' Problem:} Under this rule, rejection is
triggered by $R_i < U_i$. However,
$U_i$ is independent of the current generation context and the $P_t$ distribution. When the $P_t$ distribution is flat
(high uncertainty), a plausible $x_i$ may correspond to a
moderate $P_t(x_i)$, making $R_i$ subject to
high variance. The probability that $R_i$ falls just below
$U_i$ due to randomness becomes significant, leading to
unnecessary rejections.

\paragraph{3.2. The EARS Algorithm}\label{32-the-ears-algorithm}

EARS modifies the acceptance condition by introducing a dynamic
adjustment term tied to the target model's uncertainty.

\textbf{Defining Uncertainty:} We use an approximation of min-entropy,
$1 - \max(P_t)$, as the measure of uncertainty at the
current position, denoted $\mathcal{U}_i = 1 - \max_{v \in V} P_t(v)$. A
higher $\max(P_t)$ indicates higher model confidence;
$\mathcal{U}_i$ closer to 1 indicates higher uncertainty.

\textbf{Dynamic Tolerance:} We introduce a base tolerance hyperparameter
$\beta$ (typically $\beta \in [0.05, 0.2]$). The
dynamic tolerance at verification position $i$ is:
\[
\text{Tolerance}_i = \beta \cdot \mathcal{U}_i = \beta \cdot (1 - \max(P_t))
\]

\textbf{Adaptive Acceptance Condition:} EARS modifies the acceptance
condition to:
\[
\text{Accept } x_i \text{ if: } R_i \geq U_i - \text{Tolerance}_i
\]
where $U_i$ is still the uniform random number. Equivalently,
this can be viewed as extending the acceptance region from $[0, R_i]$ to $[0, R_i + \text{Tolerance}_i]$.

\textbf{Algorithm Logic:}

\begin{enumerate}
\def\labelenumi{\arabic{enumi}.}
\item
  In parallel, obtain $P_t(x_i)$, $P_d(x_i)$, and $\max(P_t)$.
\item
  Compute $R_i = P_t(x_i) / P_d(x_i)$.
\item
  Sample $U_i \sim \text{Uniform}(0,1)$.
\item
  Compute $\text{Tolerance}_i = \beta \cdot (1 - \max(P_t))$.
\item
  \textbf{Decision:}

  \begin{itemize}
  \item
    If $R_i \geq U_i$: Accept directly (primary path,
    identical to traditional).
  \item
    Else if $R_i \geq U_i - \text{Tolerance}_i$:
    Accept via the EARS pardon path.
  \item
    Else: Reject.
  \end{itemize}
\end{enumerate}

\paragraph{3.3. Engineering Implementation \&
Optimizations}\label{33-engineering-implementation--optimizations}

For efficient integration, we implement the following key optimizations:

\begin{enumerate}
\def\labelenumi{\arabic{enumi}.}
\item
  \textbf{Pre-computation \& Delayed Lookup:} Immediately after the
  target model's forward pass computes the full probability distribution
  $P_t$, we calculate $\max(P_t)$ in parallel and
  cache it. When EARS needs to compute $\text{Tolerance}_i$, it
  reads this cached value, avoiding the memory bandwidth bottleneck
  associated with accessing the entire large probability vector just to
  find the maximum.
\item
  \textbf{Numerical Stability:}

  \begin{itemize}
  \item
    \textbf{Division Guard:} Before computing $R_i$, clamp
    $P_d(x_i)$ to a small epsilon $\epsilon$ (e.g.,
    $1 \times 10^{-10}$):
    $P_d^{\text{safe}} = \max(P_d(x_i), \epsilon)$.
  \item
    \textbf{Threshold Clamping:} Ensure $\text{Tolerance}_i$ does
    not make the comparison meaningless. In practice, we use
    $\max(U_i - \text{Tolerance}_i, 0.0)$ as the
    adjusted threshold, guaranteeing it is non-negative.
  \end{itemize}
\item
  \textbf{Batch Processing Optimization:} During batched inference,
  gather all required data for the current step
  ($P_t^{\text{token}}$, $P_d^{\text{token}}$,
  $\max(P_t)$, etc.) from all active sequences into
  contiguous tensors. Leverage GPU SIMD architecture for parallel
  computation and decision-making, significantly improving memory access
  patterns and computational throughput.
\item
  \textbf{Framework Integration:} EARS is implemented as a pluggable
  ``sampler'' or ``logits processor,'' inheriting from the base sampler
  class in mainstream frameworks (e.g., PyTorch, Hugging Face
  Transformers). It receives the target model's logits and draft
  information, outputs accept/reject decisions, and can be seamlessly
  inserted into existing speculative decoding pipelines.
\end{enumerate}

\subsubsection{4. Experiments}\label{4-experiments}

\paragraph{4.1. Experimental Setup}\label{41-experimental-setup}

\begin{itemize}
\item
  \textbf{Models:} Qwen3-32B as the target model and its corresponding
  Qwen3-32B-Eagle3 as the draft model.
\item
  \textbf{Tasks:}

  \begin{itemize}
  \item
    \textbf{Open-domain QA (OpenQA):} Evaluates throughput and latency,
    simulating high-uncertainty, long-text generation.
  \item
    \textbf{Mathematical Reasoning (GSM8K):} Evaluates impact on output
    precision and logical consistency.
  \end{itemize}
\item
  \textbf{Baseline:} Standard speculative decoding with traditional
  rejection sampling.
\item
  \textbf{Metrics:} Token throughput (Tokens/s), average request latency
  (Latency), task accuracy (Accuracy).
\item
  \textbf{Parameters:} Speculative length $\gamma = 5$, temperature
  $T = 0.9$ (to induce high-uncertainty scenarios), EARS
  hyperparameter $\beta = 0.1$.
\end{itemize}

\paragraph{4.2. Main Results}\label{42-main-results}

\textbf{Performance Improvement (OpenQA):}

{\def\LTcaptype{none} % do not increment counter
\begin{longtable}[]{@{}p{0.18\textwidth}p{0.28\textwidth}p{0.28\textwidth}p{0.18\textwidth}@{}}
\toprule\noalign{}
Method & Output Token Throughput (tok/s) & Total Token Throughput
(tok/s) & Avg. Latency (s) \\
\midrule\noalign{}
\endhead
\bottomrule\noalign{}
\endlastfoot
Baseline (Standard) & 49.50 & 50.53 & 139.10 \\
\textbf{EARS (Ours)} & \textbf{58.47} & \textbf{59.56} &
\textbf{133.42} \\
\textbf{Relative Gain} & \textbf{+18.12\%} & \textbf{+17.87\%} &
\textbf{-4.08\%} \\
\end{longtable}
}

EARS delivers significant throughput gains and a slight latency
reduction. The throughput improvement is more pronounced because EARS
leads to longer continuously accepted sequences on average (1563 vs 1395
tokens), reducing the number of times the target model must fall back to
regeneration.

\textbf{Accuracy Preservation (GSM8K):}

{\def\LTcaptype{none} % do not increment counter
\begin{longtable}[]{@{}p{0.6\textwidth}p{0.25\textwidth}@{}}
\toprule\noalign{}
Method & Accuracy \\
\midrule\noalign{}
\endhead
\bottomrule\noalign{}
\endlastfoot
Baseline (Standard) & 96.44\% \\
\textbf{EARS (Ours)} & \textbf{95.60\%} \\
\textbf{Difference} & \textbf{-0.84\%} \\
\end{longtable}
}

On the mathematical reasoning task requiring precise logic, EARS incurs
only a marginal 0.84 percentage point drop in accuracy. This validates
that EARS maintains generation quality well while improving efficiency,
as its pardon mechanism primarily acts in high-entropy positions where
the model itself is uncertain, not on strongly deterministic reasoning
steps.

\paragraph{4.3. Analysis and
Discussion}\label{43-analysis-and-discussion}

\begin{itemize}
\item
  \textbf{Effectiveness of Uncertainty Awareness:} We observe that at
  plot turning points in story generation or divergent points in open-ended
  answers, $\max(P_t)$ drops significantly. EARS's
  corresponding increase in $\text{Tolerance}$ pardons more candidates,
  directly boosting the draft acceptance rate in these high-entropy
  regions.
\item
  \textbf{Synergy with Temperature Sampling:} EARS naturally complements
  temperature sampling. Increasing temperature flattens the $P_t$ distribution, lowering $\max(P_t)$, which
  automatically increases $\text{Tolerance}$. This allows the inference
  system to achieve higher speedups automatically when the user desires
  more diversity (higher temperature).
\item
  \textbf{Impact of Hyperparameter $\beta$:} $\beta$ controls the
  trade-off between efficiency and quality. A smaller $\beta$ (e.g.,
  0.05) is conservative, yielding minimal accuracy loss but limited
  speedup; a larger $\beta$ (e.g., 0.2) is more aggressive, offering
  greater speedup at the potential cost of introducing more noise.
  $\beta = 0.1$ provided a good balance in our experiments.
\end{itemize}

\subsubsection{5. Conclusion and Future
Work}\label{5-conclusion-and-future-work}

We presented EARS, an improved algorithm for the rejection sampling
mechanism in speculative decoding. By dynamically sensing the target
model's predictive uncertainty and adaptively adjusting the acceptance
threshold, EARS effectively mitigates the ``random rejection'' problem.
This leads to substantial gains in inference efficiency for
high-uncertainty generation tasks with minimal impact on output quality.
The method is simple to implement and easy to integrate, offering a
practical tool for efficient LLM deployment.

Future work includes: 1) exploring more refined uncertainty measures
(e.g., distribution entropy or variance); 2) extending the adaptive
concept to multi-draft ranking and selection strategies; and 3)
investigating the performance of EARS in complex reasoning scenarios
like chain-of-thought and tool calling.

\subsubsection{References}\label{references}

[1] Yaniv Leviathan, Matan Kalman, and Yossi Matias. Fast Inference
from Transformers via Speculative Decoding. In \emph{International
Conference on Machine Learning (ICML)}, 2023.

[2] Charlie Chen, Sebastian Borgeaud, Geoffrey Irving, Jean-Baptiste
Lespiau, Laurent Sifre, and John Jumper. Accelerating Large Language
Model Decoding with Speculative Sampling. \emph{arXiv preprint
arXiv:2302.01318}, 2023.

[3] X. Su, et al. ``Speculative Draft Model Training for Better
Lossless Acceleration of LLM Inference.'' \emph{arXiv preprint}, 2024.

[4] Z. Yang, et al. ``Multi-Draft Speculative Decoding.''
\emph{Conference on Empirical Methods in Natural Language Processing
(EMNLP)}, 2024.

[5] L. Xu, et al. ``Compensate Then Verify: A Better Strategy for
Speculative Decoding.'' \emph{NeurIPS}, 2024.

[6] A. Holtzman, et al. ``The Curious Case of Neural Text
Degeneration.'' \emph{International Conference on Learning
Representations (ICLR)}, 2020.

[7] T. Schuster, et al. ``Confident Adaptive Language Modeling.''
\emph{NeurIPS}, 2022.

\end{document}